\documentclass[a4paper,fleqn]{cas-dc}

\usepackage[authoryear,longnamesfirst]{natbib}
\usepackage{subcaption}
\usepackage{todonotes}

\def\tsc#1{\csdef{#1}{\textsc{\lowercase{#1}}\xspace}}
\tsc{WGM}
\tsc{QE}
\tsc{EP}
\tsc{PMS}
\tsc{BEC}
\tsc{DE}

\begin{document}
\let\WriteBookmarks\relax
\def\floatpagepagefraction{1}
\def\textpagefraction{.001}

\shorttitle{Validation of an AI TUNEL Assay}

\shortauthors{B. A. Jacobs, A. Morris I. Shaik, F. Lin}

\title [mode = title]{Validation of an Artificial Intelligence Tool for the Detection of Sperm DNA Fragmentation Using the TUNEL In Situ Hybridization Assay}



%
\author[1,3]{B. A. Jacobs}[orcid = 0000-0002-3121-3565]
\ead{byron.jacobs@vitruvianmd.com}
\credit{Conceptualization of this study, Computational Methodology, Algorithm Design, Writing}
\affiliation[1]{organization={University of Johannesburg},
    addressline={Department of Mathematics and Applied Mathematics, Auckland Park, PO Box 524}, 
    city={Johannesburg},
    postcode={2006}, 
    state={Gauteng},
    country={South Africa}}
\cormark[1]

\author[2]{A. Morris}


\ead{aqeel.morris@up.ac.za}


\credit{Biological Methodology, Writing}

\affiliation[2]{organization={University Pretoria},
    addressline={Faculty of Health Sciences, Department of Obstetrics and Gynaecology, Steve Biko Academic Hospital, Hatfield, Private Bag X20}, 
    city={Pretoria},
    postcode={0028}, 
    state={Gauteng},
    country={South Africa}}

\author[3]{I. Shaik}
\credit{Conceptualization of this study, editing}
\author[3]{F. Lin}
\credit{Conceptualization of this study, editing}

\affiliation[3]{organization={VitruvianMD Pte Ltd},
    addressline={3 Fraser Street}, 
    city={Singapore},
    postcode={189352}, 
    country={Singapore}}
    
\cortext[cor1]{Corresponding author}



\begin{abstract}
Sperm DNA fragmentation (SDF) is a critical parameter in male fertility assessment that conventional semen analysis fails to evaluate. This study presents the validation of a novel artificial intelligence (AI) tool designed to detect SDF through digital analysis of phase contrast microscopy images, using the terminal deoxynucleotidyl transferase dUTP nick end labeling (TUNEL) assay as the gold standard reference. Utilising the established link between sperm morphology and DNA integrity, the present work proposes a morphology assisted ensemble AI model that combines image processing techniques with state-of-the-art transformer based machine learning models (GC-ViT) for the prediction of DNA fragmentation in sperm from phase contrast images. The ensemble model is benchmarked against a pure transformer `vision' model as well as a `morphology-only` model. Promising results show the proposed framework is able to achieve sensitivity of 60\% and specificity of 75\%. This non-destructive methodology represents a significant advancement in reproductive medicine by enabling real-time sperm selection based on DNA integrity for clinical diagnostic and therapeutic applications.
\end{abstract}


\begin{highlights}
\item A novel ensemble model, comprising both machine learning and image processing techniques.
\item Digitisation of the TUNEL assay allows for the analysis of intra-expert variance
\item The proposed model is non-destructive, allowing for the selection of viable sperm for use in ARTs.
\end{highlights}

\maketitle
\section{Introduction}
Infertility affects approximately 15\% of couples of childbearing age worldwide, with male factors contributing to 30-50\% of these cases \cite{mascarenhas2012national, agarwal2015unique}. While conventional semen analysis remains the cornerstone of male fertility assessment, it fails to evaluate a critical parameter: sperm DNA fragmentation (SDF). SDF refers to breaks in single or double-stranded DNA within the sperm nucleus and has been associated with reduced fertilization rates, impaired embryo development, increased miscarriage rates, and potentially adverse health outcomes in offspring \cite{henkel2007dna}.\\ \\
The terminal deoxynucleotidyl transferase dUTP nick end labeling (TUNEL) assay has emerged as one of the most reliable methods for detecting SDF. This technique identifies DNA strand breaks by labeling the terminal end of nucleic acids, providing a quantitative assessment of SDF that correlates well with clinical outcomes \cite{hamidi2015double}. The integrity of DNA is directly measured by detection of single and double strand breaks through an enzymatic reaction that labels the free 3'-OH termini with modified nucleotides via terminal deoxynucleotidyl transferase. Sperm with intact DNA show minimal background staining (TUNEL-negative), while those with fragmented DNA exhibit bright fluorescence (TUNEL-positive) \cite{world2021laboratory}.\\ \\
Despite its clinical utility, the TUNEL assay presents several limitations that hinder its widespread implementation. The technique requires specialized equipment, trained personnel, and is time-consuming. Moreover, the fixation and staining procedures render the assessed sperm non-viable, preventing their subsequent use in assisted reproductive technologies (ARTs) such as in vitro fertilization (IVF) and intracytoplasmic sperm injection (ICSI) \cite{henkel2007dna}. This represents a significant drawback in fertility treatment scenarios where the identification and selection of sperm with intact DNA would be highly beneficial for clinical outcomes.\\ \\
The field of sperm DNA fragmentation assessment is further complicated by the variety of available assays, including Acridine Orange Test (AOT), Chromomycin A3 (CMA3), Sperm Chromatin Structure Assay (SCSA), Single Cell Gel Electrophoresis (COMET), and Sperm Chromatin Dispersion (SCD). These tests measure different aspects of DNA damage and chromatin integrity, often yielding inconsistent results \cite{perez2011comparison}. \cite{hamidi2015double} reported "perfectly comparable" results between TUNEL and AOT, while other studies have found discrepancies between techniques and their clinical relevance \cite{nijs2009chromomycin, ghasemzadeh2015sperm}. This inconsistency highlights the need for a standardized, reliable approach to SDF assessment.\\ \\
Recent advances in artificial intelligence (AI) and machine learning have revolutionized various aspects of healthcare diagnostics, including reproductive medicine \cite{wang2019artificial}. Neural networks, in particular, have demonstrated remarkable capabilities in image recognition and classification tasks that were previously dependent on human expertise. Several studies have explored the application of AI in sperm analysis with promising results. \cite{serrano2022p} used random forests and convolutional neural networks to predict the fragmentation results of COMET with high precision. \cite{wang2019artificial} developed linear and non-linear regression models trained on data from the AOT, achieving a test accuracy of 82.7\%. However, these approaches often focus on a single assay and may not provide a comprehensive assessment of sperm DNA integrity.\\ \\
The development of an AI-based tool that can accurately detect SDF without compromising sperm viability would represent a significant advance in reproductive medicine. Such a system could potentially identify subtle morphological or biochemical signatures associated with DNA fragmentation that are imperceptible to human observers, while maintaining the functional capacity of the sperm for subsequent ART procedures.\\ \\
This study presents the validation of a novel AI tool designed to detect sperm DNA fragmentation using the TUNEL in situ hybridization assay as the gold standard reference. Our approach differs from previous studies in several key aspects. First, we employ a machine learning framework that digitally replicates chemical tests using phase-contrast microscopy images alone, eliminating the need for destructive chemical assays. Second, our methodology incorporates morphological parameters as metadata to enhance prediction accuracy. Finally, our system is amenable to the processing of both mobile and immobilized spermatozoa, whether alive or dead, making it particularly valuable for sperm selection in IVF or ICSI procedures. This non-invasive, efficient approach has the potential to significantly improve ART outcomes by ensuring that only sperm with intact DNA integrity are selected for use. \\ \\
In addition to the novel ensemble approach to sperm classification, this paper presents a novel dataset from which we are able to quantify intra-expert variance across a cohort of patients. This variability informs the bounds of performance of an AI assitive tool. Finally, the digitisation of these kinds of data allows for a rich analysis of inter-expert variance, which is the subject of ongoing work.

\section{Materials and Methods}
\subsection{Sample Collection and Preparation}
Semen samples were collected from 35 consenting patients following standard laboratory procedures. Inclusion criteria encompassed parameters above the lower reference limits as stipulated by the WHO laboratory manual for the examination and processing of human semen, Sixth Edition. Samples indicating possible genital tract infection with increased leukocytes (greater than $1.0 \times 10^6$ cells/ml for peroxidase-positive cells) were included. Exclusion criteria comprised azoospermia, high viscosity, and poor liquefaction.

\subsection{TUNEL Assay}
The TUNEL assay was performed according to the manufacturer's instructions for ApopTag Plus Peroxidase in situ apoptosis detection kit (Merck, Kenilworth, NJ, USA). The integrity of the DNA was directly measured by detection of single and double strand breaks. These DNA strand breaks were detected in an enzymatic reaction by labeling the free 3'-OH termini with modified nucleotides via terminal deoxynucleotidyl transferase. Sperm cells with intact DNA showed slight background staining (TUNEL-negative), while those with fragmented DNA exhibited bright green fluorescence (TUNEL-positive).

\subsection{Image Acquisition}
The prepared semen smear slides were viewed and digitally imaged using VitruvianMD's VisionMD camera and accompanying image capture suite. Different fields of view were imaged under phase contrast, bright field, and fluorescence until a minimum of 100 spermatozoa per patient were observed. In cases where this target was not achievable due to low sperm count, all available spermatozoa were imaged.

\subsection{Data Description}
The dataset comprised 1825 image triples (bright-field, phase-contrast, and fluorescence) of individual spermatozoa. The images collected were taken from 35 consenting patients whose samples were prepared for the TUNEL assay according to standard protocol. Of these cells, 512 were identified as fragmented, 715 as not fragmented, and 591 received a `null` annotation where the expert could not reliably classify the fluorescent image. This highlights the subjective nature of current assessment methods and underscores the need for a quantitative, reliable, and repeatable solution.\\ \\
Sample images from the three different annotation classes (fragmented, unfragmented and null) are presented in Appendix \ref{sec:data_example}, Figure \ref{fig:sample_grid}. Each of these sperm were imaged under bright-field, phase-contrast and fluorescent lighting conditions. These example images indicate the importance of expert annotation as the null labelled sperm exhibits some degree of fluorescence. However, this also underpins the uncertainty associated with inter-expert variance and highlights the need for a quantitative, reliable and repeatable solution. In the present study, all images that received a `null` annotation have been excluded from the dataset.\\ \\
    
\subsection{Intra-expert variance}\label{sec:intra_expert_variance}
The fluorescent images were annotated by a single expert on two separate occasions, ten months apart. The purpose of this was to assess the intra-expert variance and consistency of the annotations. The expert was blinded to their previous annotations during the second round of labeling. It was found that on a per sperm basis, the expert annotations agreed on 81\% of the images. This indicates a reasonable level of consistency, although it also highlights the inherent subjectivity in the manual assessment of sperm DNA fragmentation. In addition to the per-sperm analysis, we also analysed the outcome of the TUNEL assay on a per-patient basis. The outcome of the TUNEL assay is reported as the percentage of sperm with fragmented DNA, counting 200 spermatozoa per sample. It was found that the difference in reported SDF \% had an absolute mean difference of 13.7\% and standard deviation of 19.5\%, further highlighting the need for an objective and quantitative method for assessing sperm DNA integrity.
	
\subsection{AI Model Development}
\begin{figure}
        \centering
        \includegraphics[width=0.5\textwidth]{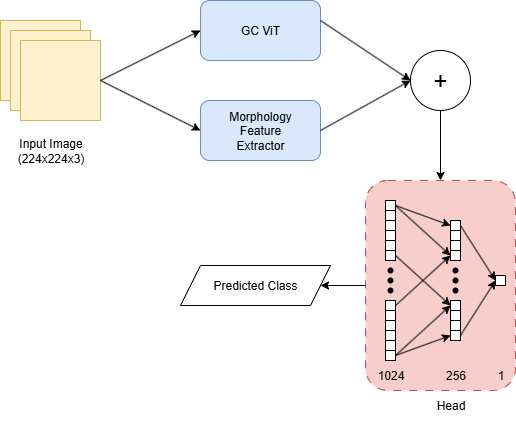}
        \caption{Model Architecture}
        \label{fig:model_architecture}
    \end{figure}

Due to the limited dataset size, transfer learning techniques were employed to enhance model performance. To avoid possible data leakage, images from the same patient were grouped together and assigned to either the training or validation set. Ultimately, the training set comprised 1017 images from 28 patients, while the validation set included 210 images from 7 patients. \\ \\
Three candidate models were architected and benchmarked. The first model is a pure vision model based on the GC-ViT transformer architecture \cite{hatamizadeh2023global}. This model processes only the phase-contrast images of individual spermatozoa. The second model is a morphology-only model that utilises morphological parameters inferred from the phase-contrast images as metadata. Among these metadata are morphological features typically assocaited with sperm morphology such as head length, head width, vacuole presesnce and acrosomal area, measured using the technique established in \cite{jacobs2024image}. Finally, we propose an ensemble model that combines the extracted morphological features with the base transformer model. This ensemble approach leverages the strengths of both image-based and morphology-based features to enhance predictive accuracy.\\ \\
Figure \ref{fig:model_architecture} illustrates the model architectures, where the additive module selects for either the vision features, morphological features, or a combination of both. The head architecture was tuned and found that two layers with 1024 and 256 nodes and leaky rectified linear units activation functions acheived best results, with dropout rates of 0.6 and 0.3 employed as a regularization strategy. \\ \\
All the models were trained using a binary cross-entropy loss function and the Adam optimizer with an initial learning rate of $5\times10^{-5}$. A layer-wise learning rate decay strategy was implemented, with a decay factor of 0.12 applied to each successive layer as well as a warmup proportion of 0.1. This approach helps to fine-tune the pre-trained weights of the transformer model while preventing overfitting on the limited dataset. Finally, early stopping was implemented to prevent overfitting, with training ceasing if the validation loss did not improve for 10 consecutive epochs. Data augmentation techniques, including random rotations and flips, were applied to the training images to enhance model generalization. The models were trained for a maximum of 50 epochs, with the best-performing model on the validation set selected for final evaluation.\\ \\ 

\section{Results}
\begin{figure}
        \centering
        \includegraphics[width=0.5\textwidth]{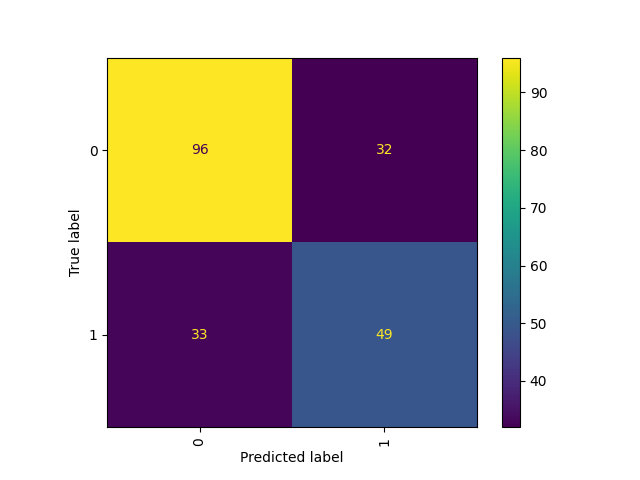}
        \caption{Ensemble Model Confusion Matrix}
        \label{fig:ensemble_confusion_matrix}
    \end{figure}

\begin{figure}
        \centering
        \includegraphics[width=0.5\textwidth]{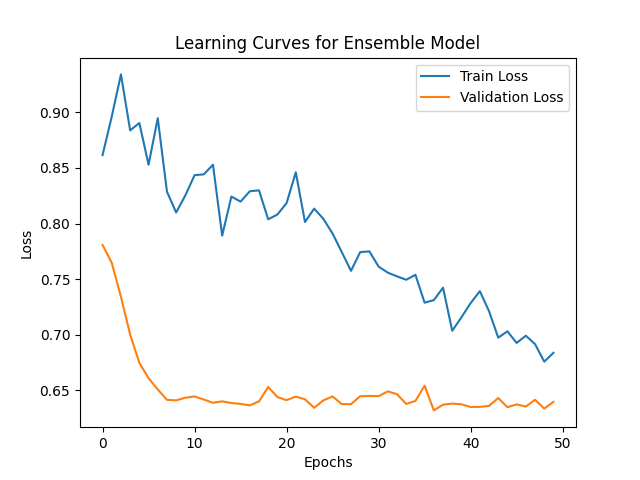}
        \caption{Ensemble Model Learning Curves}
        \label{fig:ensemble_learning_curves}
    \end{figure}

\begin{figure}
        \centering
        \includegraphics[width=0.5\textwidth]{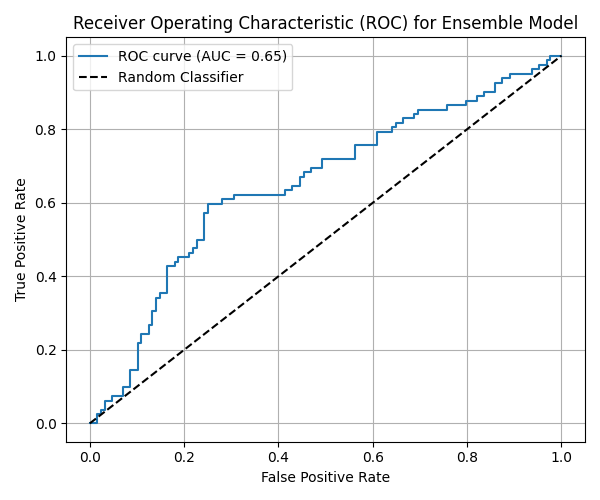}
        \caption{Ensemble Model ROC Curve}
        \label{fig:ensemble_roc_curve}
    \end{figure}

\begin{figure}
        \centering
        \includegraphics[width=0.5\textwidth]{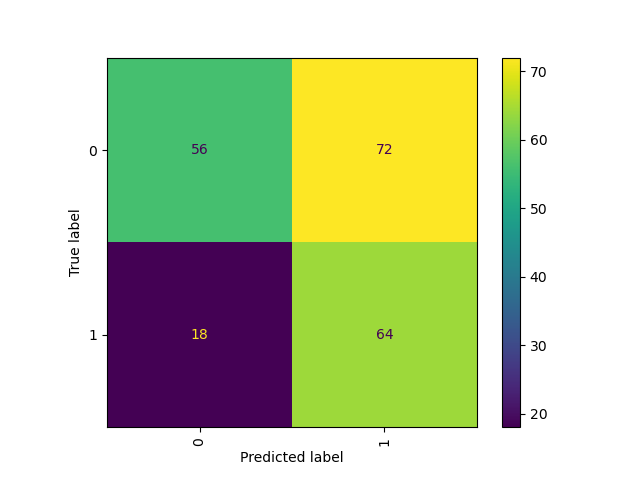}
        \caption{Morphology Model Confusion Matrix}
        \label{fig:morph_confusion_matrix}
    \end{figure}

\begin{figure}
        \centering
        \includegraphics[width=0.5\textwidth]{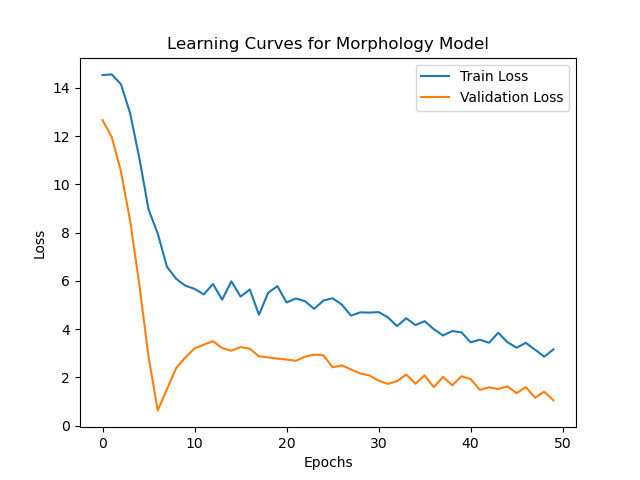}
        \caption{Morphology Model Learning Curves}
        \label{fig:morph_learning_curves}
    \end{figure}

\begin{figure}
        \centering
        \includegraphics[width=0.5\textwidth]{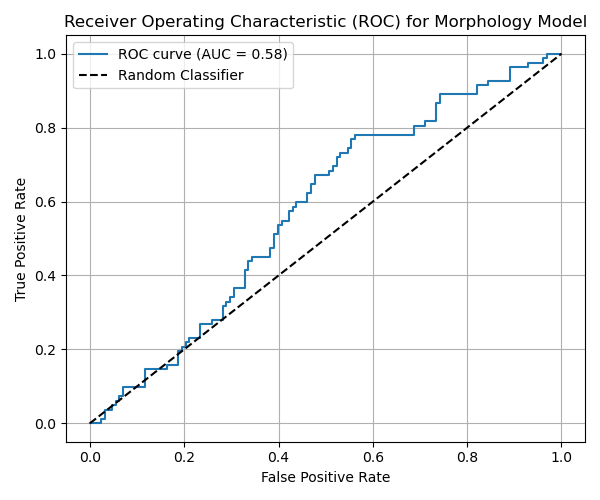}
        \caption{Morphology Model ROC Curve}
        \label{fig:morph_roc_curve}
    \end{figure}

\begin{figure}
        \centering
        \includegraphics[width=0.5\textwidth]{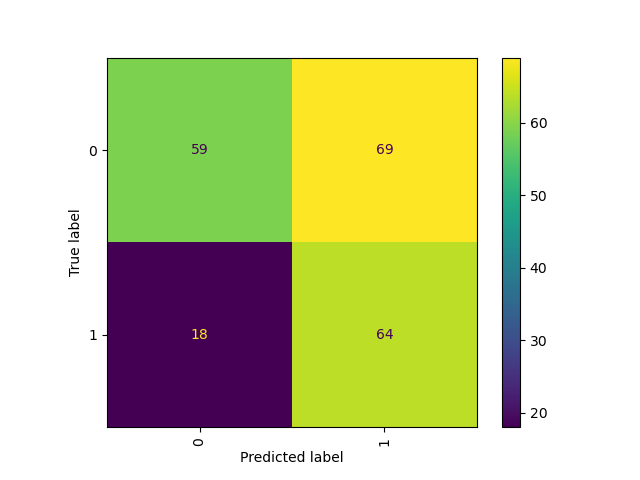}
        \caption{Vision Model Confusion Matrix}
        \label{fig:vision_confusion_matrix}
    \end{figure}

\begin{figure}
        \centering
        \includegraphics[width=0.5\textwidth]{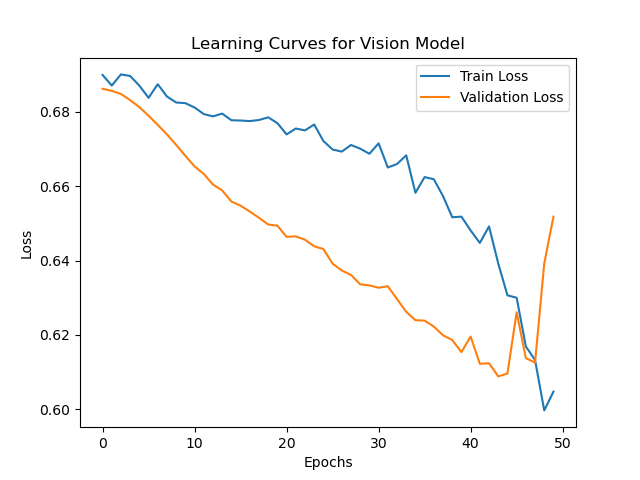}
        \caption{Vision Model Learning Curves}
        \label{fig:vision_learning_curves}
    \end{figure}

\begin{figure}
        \centering
        \includegraphics[width=0.5\textwidth]{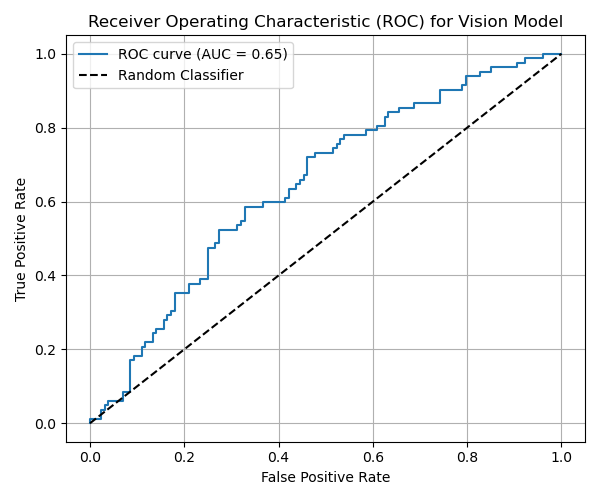}
        \caption{Vision Model ROC Curve}
        \label{fig:vision_roc_curve}
    \end{figure}

The performance of all three models was evaluated using standard classification metrics, including sensitivity (recall), specificity, precision, accuracy, and the $F_1$ score. The results are summarized in the classification reports, Tables \ref{tab:ensemble_classification_report}, \ref{tab:morphology_classification_report}, and \ref{tab:vision_classification_report} for the Ensemble model, Morphology model, and Vision model, respectively. We note that the precision and recall of the Ensemble model achieves a good balance of predict both the fragmented and unfragmented classes. Perhaps this is more clearly illustrated in the confusion matrices; Figures \ref{fig:ensemble_confusion_matrix}, \ref{fig:morph_confusion_matrix}, and \ref{fig:vision_confusion_matrix}, where the Vision and Morphology models bias toward predicting the fragmented class.\\ \\
The performance metrics and confidence intervals are presented in Tables \ref{tab:model_performance_vision_morphology}, \ref{tab:model_performance_morphology}, and \ref{tab:model_performance_vision}. These results again highlight the balanced and overall better performance of the Ensemble model. \\ \\
Finally, all three Recievier Operating Characteristic (ROC) curves, presented in Figures \ref{fig:ensemble_roc_curve}, \ref{fig:morph_roc_curve}, and \ref{fig:vision_roc_curve}, illustrate a clear advantage over random classification, with the Ensemble model perfoming the best.\\ \\
We note that the learning curves, Figures \ref{fig:ensemble_learning_curves}, \ref{fig:morph_learning_curves}, and \ref{fig:vision_learning_curves}, indicate that the models are not overfitting, with the exception of the Vision model which just begins to show signs of overfitting in the final few epochs. This is mitigated by the early stopping criteria ensuring that the model with the lowest validation loss is preserved for the performance benchmarks. Training and validation accuracy suggests that additional training data could further enhance model performance, however due to the cost of data collection and expert time required for annotation, additional data are not feasible to collect.\\ \\
The presented results indicate the potential of an AI driven assitive tool for the quantification of sperm analysis, and potentially sperm selection. In contrast to the intra-expert variance reported in Section \ref{sec:intra_expert_variance}, the proposed models are objective and quantitative. Additionally, the morphological and enesmble model enjoy the benefit of interpretability, as the morphological features can be directly linked to the classification outcome. In order to directly compare the perfomance of the ensemeble model (or other computer assisted models), the inter-expert variance must be quantified. This would help to contextualise the true accuracy of the propsoed models, rather than the alignment of the model to a single expert.

\begin{table}[h!]
\centering
\caption{Ensemble Model Classification Report}
\label{tab:ensemble_classification_report}
\resizebox{0.5 \textwidth}{!}{%
\begin{tabular}{|l|c|c|c|c|}
\hline
\textbf{Class} & \textbf{Precision} & \textbf{Recall} & \textbf{$F_1$-score} & \textbf{Support} \\
\hline
Unfragmented (0) & 0.74 & 0.75 & 0.75 & 128 \\
Fragmented (1) & 0.60 & 0.60 & 0.60 & 82 \\
\hline
accuracy & & & 0.69 & 210 \\
macro avg & 0.67 & 0.67 & 0.67 & 210 \\
weighted avg & 0.69 & 0.69 & 0.69 & 210 \\
\hline
\end{tabular}%
}
\end{table}

\begin{table}[h!]
\centering
\caption{Morphology Model Classification Report}
\label{tab:morphology_classification_report}
\resizebox{0.5 \textwidth}{!}{%
\begin{tabular}{|l|c|c|c|c|}
\hline
\textbf{Class} & \textbf{Precision} & \textbf{Recall} & \textbf{$F_1$-score} & \textbf{Support} \\
\hline
Unfragmented (0) & 0.76 & 0.44 & 0.55 & 128 \\
Fragmented (1) & 0.47 & 0.78 & 0.59 & 82 \\
\hline
accuracy & & & 0.57 & 210 \\
macro avg & 0.61 & 0.61 & 0.57 & 210 \\
weighted avg & 0.65 & 0.57 & 0.57 & 210 \\
\hline
\end{tabular}%
}
\end{table}

\begin{table}[h!]
\centering
\caption{Vision Model Classification Report}
\label{tab:vision_classification_report}
\resizebox{0.5 \textwidth}{!}{%
\begin{tabular}{|l|c|c|c|c|}
\hline
\textbf{Class} & \textbf{Precision} & \textbf{Recall} & \textbf{$F_1$-score} & \textbf{Support} \\
\hline
Unfragmented (0) & 0.77 & 0.46 & 0.58 & 128 \\
Fragmented (1)& 0.48 & 0.78 & 0.60 & 82 \\
\hline
accuracy & & & 0.59 & 210 \\
macro avg & 0.62 & 0.62 & 0.59 & 210 \\
weighted avg & 0.65 & 0.59 & 0.58 & 210 \\
\hline
\end{tabular}%
}
\end{table}

    \begin{table}[h!]
    \centering
    \caption{Performance Metrics for Ensemble Model}
    \label{tab:model_performance_vision_morphology}
    \begin{tabular}{|l|c|}
    \hline
    \textbf{Metric} & \textbf{Value (95\% CI)} \\
    \hline
    Sensitivity (Recall) & 0.60 (0.48 - 0.70) \\
    Specificity & 0.75 (0.67 - 0.82) \\
    Precision & 0.60 (0.49 - 0.71) \\
    Accuracy & 0.69 (0.62 - 0.75) \\
    F1-Score & 0.60 \\
    \hline
    \end{tabular}
    \end{table}

    \begin{table}[h!]
    \centering
    \caption{Performance Metrics for Morphology Model}
    \label{tab:model_performance_morphology}
    \begin{tabular}{|l|c|}
    \hline
    \textbf{Metric} & \textbf{Value (95\% CI)} \\
    \hline
    Sensitivity (Recall) & 0.76 (0.64 - 0.85) \\
    Specificity & 0.44 (0.35 - 0.53) \\
    Precision & 0.44 (0.35 - 0.53) \\
    Accuracy & 0.55 (0.48 - 0.62) \\
    F1-Score & 0.55 \\
    \hline
    \end{tabular}
    \end{table}

    \begin{table}[h!]
    \centering
    \caption{Performance Metrics for Vision Model}
    \label{tab:model_performance_vision}
    \begin{tabular}{|l|c|}
    \hline
    \textbf{Metric} & \textbf{Value (95\% CI)} \\
    \hline
    Sensitivity (Recall) & 0.87 (0.77 - 0.93) \\
    Specificity & 0.39 (0.31 - 0.48) \\
    Precision & 0.48 (0.39 - 0.56) \\
    Accuracy & 0.58 (0.51 - 0.64) \\
    F1-Score & 0.61 \\
    \hline
    \end{tabular}
    \end{table}

\section{Classification Examples}
\begin{figure*}[t!]
    \centering
    \begin{subfigure}[b]{0.25\textwidth}
        \centering
        \includegraphics[width=\textwidth]{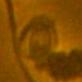}
        \caption{Example 1:\\ True Label - Fragmented,\\ Ensemble - Fragmented,\\ Vision - Fragmented,\\ Morphology - Fragmented}
        \label{fig:ex1}
    \end{subfigure}
    \hfill
    \begin{subfigure}[b]{0.25\textwidth}
        \centering
        \includegraphics[width=\textwidth]{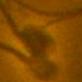}
        \caption{Example 2:\\ True Label - Fragmented,\\ Ensemble - Fragmented,\\ Vision - Fragmented,\\ Morphology - Unfragmented}
        \label{fig:ex2}
    \end{subfigure}
    \hfill
    \begin{subfigure}[b]{0.25\textwidth}
        \centering
        \includegraphics[width=\textwidth]{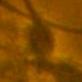}
        \caption{Example 3:\\ True Label - Fragmented,\\ Ensemble - Fragmented,\\ Vision - Unfragmented,\\ Morphology - Fragmented}
        \label{fig:ex3}
    \end{subfigure}
    
    \vspace{1em} 

    \begin{subfigure}[b]{0.25\textwidth}
        \centering
        \includegraphics[width=\textwidth]{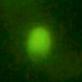}
        \caption{Example 1: Fluorescent Image}
        \label{fig:ex1fl}
    \end{subfigure}
    \hfill
    \begin{subfigure}[b]{0.25\textwidth}
        \centering
        \includegraphics[width=\textwidth]{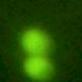}
        \caption{Example 2: Fluorescent Image}
        \label{fig:ex2fl}
    \end{subfigure}
    \hfill
    \begin{subfigure}[b]{0.25\textwidth}
        \centering
        \includegraphics[width=\textwidth]{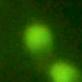}
        \caption{Example 3: Fluorescent Image}
        \label{fig:ex3fl}
    \end{subfigure}

    \vspace{1em} 

    \begin{subfigure}[b]{0.25\textwidth}
        \centering
        \includegraphics[width=\textwidth]{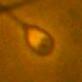}
        \caption{Example 4:\\ True Label - Unfragmented,\\ Ensemble - Unfragmented,\\ Vision - Unfragmented,\\ Morphology - Unfragmented}
        \label{fig:ex4}
    \end{subfigure}
    \hfill
    \begin{subfigure}[b]{0.25\textwidth}
        \centering
        \includegraphics[width=\textwidth]{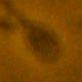}
        \caption{Example 5:\\ True Label - Unfragmented,\\ Ensemble - Unfragmented,\\ Vision - Fragmented,\\ Morphology - Unfragmented}
        \label{fig:ex5}
    \end{subfigure}
    \hfill
    \begin{subfigure}[b]{0.25\textwidth}
        \centering
        \includegraphics[width=\textwidth]{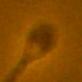}
        \caption{Example 6:\\ True Label - Unfragmented,\\ Ensemble - Unfragmented,\\ Vision - Fragmented,\\ Morphology - Fragmented}
        \label{fig:ex6}
    \end{subfigure}

    \vspace{1em} 

    \begin{subfigure}[b]{0.25\textwidth}
        \centering
        \includegraphics[width=\textwidth]{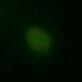}
        \caption{Example 4: Fluorescent Image}
        \label{fig:ex4fl}
    \end{subfigure}
    \hfill
    \begin{subfigure}[b]{0.25\textwidth}
        \centering
        \includegraphics[width=\textwidth]{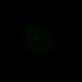}
        \caption{Example 5: Fluorescent Image}
        \label{fig:ex5fl}
    \end{subfigure}
    \hfill
    \begin{subfigure}[b]{0.25\textwidth}
        \centering
        \includegraphics[width=\textwidth]{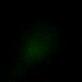}
        \caption{Example 6: Fluorescent Image}
        \label{fig:ex6fl}
    \end{subfigure}

    \caption{Array of example images correctly classified by the Ensemble model.}
    \label{fig:correct_example_grid}
\end{figure*}

\begin{figure}[t!]
    \centering
    \begin{subfigure}[b]{0.22\textwidth}
        \centering
        \includegraphics[width=\textwidth]{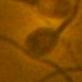}
        \caption{Example 7:\\ True Label - Fragmented,\\ Ensemble - Unfragmented,\\ Vision - Fragmented,\\ Morphology - Unfragmented}
        \label{fig:ex7}
    \end{subfigure}
    \hfill
    \begin{subfigure}[b]{0.22\textwidth}
        \centering
        \includegraphics[width=\textwidth]{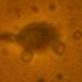}
        \caption{Example 8:\\ True Label - Fragmented,\\ Ensemble - Unfragmented,\\ Vision - Unfragmented,\\ Morphology - Fragmented}
        \label{fig:ex8}
    \end{subfigure}
    \vspace{1em} 
    \begin{subfigure}[b]{0.22\textwidth}
        \centering
        \includegraphics[width=\textwidth]{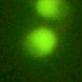}
        \caption{Example 7: Fluorescent Image}
        \label{fig:ex7fl}
    \end{subfigure}
    \hfill
    \begin{subfigure}[b]{0.22\textwidth}
        \centering
        \includegraphics[width=\textwidth]{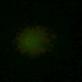}
        \caption{Example 8: Fluorescent Image}
        \label{fig:ex8fl}
    \end{subfigure}
    \vfill

    \vspace{1em} 

    \begin{subfigure}[b]{0.22\textwidth}
        \centering
        \includegraphics[width=\textwidth]{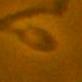}
        \caption{Example 9:\\ True Label - Unfragmented,\\ Ensemble - Fragmented,\\ Vision - Fragmented,\\ Morphology - Unfragmented}
        \label{fig:ex9}
    \end{subfigure}
    \hfill
    \begin{subfigure}[b]{0.22\textwidth}
        \centering
        \includegraphics[width=\textwidth]{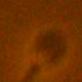}
        \caption{Example 10:\\ True Label - Unfragmented,\\ Ensemble - Fragmented,\\ Vision - Unfragmented,\\ Morphology - Fragmented}
        \label{fig:ex10}
    \end{subfigure}
    \vspace{1em} 

    \begin{subfigure}[b]{0.22\textwidth}
        \centering
        \includegraphics[width=\textwidth]{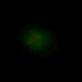}
        \caption{Example 9: Fluorescent Image}
        \label{fig:ex9fl}
    \end{subfigure}
    \hfill
    \begin{subfigure}[b]{0.22\textwidth}
        \centering
        \includegraphics[width=\textwidth]{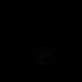}
        \caption{Example 10: Fluorescent Image}
        \label{fig:ex10fl}
    \end{subfigure}

    \caption{Array of example images incorrectly classified by the Ensemble model.}
    \label{fig:incorrect_example_grid}
\end{figure}

In this section we present example outputs from the three models discussed above. The three models were presented the phase-contrast images from the validation set. Each of the models' classifications are recorded alongside the true label in the caption of Figures \ref{fig:correct_example_grid} and \ref{fig:incorrect_example_grid}.\\ \\
The examples presented in Figure \ref{fig:incorrect_example_grid} are all misclassified by the ensemble model. Figure \ref{fig:ex7} is potentially misclassified due to the presence additional tails, leading to erroneous morphological measurements. In this case the morphological features may have biased the ensemble model the unfragmented classification.\\ \\ 
All example shown in Figure \ref{fig:correct_example_grid} are correctly classified by the ensemble model. Figures \ref{fig:ex2} illustrates the case where the input image contains extraneous information which can lead to erroneous morphological measurements. However, Figure \ref{fig:ex5} illustrates the opposite, where the input image is slightly blurry, perhaps leading to confusion in the vision model, while the sperm morphology is clearly intact. Both of these cases, the ensemble model balances the information from both sources to make a correct classification.\\ \\

\section{Discussion}
This study presents a novel approach to sperm DNA fragmentation assessment using artificial intelligence. Our machine learning-based framework provides a non-destructive method for evaluating sperm DNA integrity, addressing a critical limitation of conventional assays like TUNEL that render sperm non-viable for subsequent use in assisted reproductive technologies.\\ \\
The 75\% specificity achieved by our model represents a substantial advancement in the field of reproductive medicine. This level of performance enables reliable identification of sperm with intact DNA, which is directly correlated with improved fertilization rates, enhanced embryo viability, and higher pregnancy success rates. While the current sensitivity of 60\% leaves room for improvement, it still offers significant advantages over random selection, particularly in cases of severe male factor infertility where the proportion of normal sperm is low.\\ \\ 
Our approach builds upon previous research in several important ways. Unlike studies that focus on a single assay, our model is designed with the capability to integrate insights from multiple sperm DNA fragmentation assessment methods. Additionally, by incorporating morphological data as metadata, we enhance classification accuracy beyond what is possible with image analysis alone.\\ \\
The ability to perform rapid, non-destructive sperm assessment represents a significant leap forward in fertility treatments. By eliminating the need for expensive and destructive chemical assays, our method enables real-time sperm selection at the point of IVF/ICSI. This has the potential to not only improve ART outcomes but also provide a more accessible and cost-effective solution for couples experiencing male-factor infertility.\\ \\
It is important to acknowledge certain limitations of our study. The relatively small dataset size necessitated the use of transfer learning techniques. While this approach yielded promising results, the incorporation of additional training data is expected to further refine predictive performance, increasing both precision and specificity in future iterations. Furthermore, the model's current performance metrics, while encouraging, highlight the need for continued refinement before widespread clinical implementation.
\section{Conclusion}
This study validates a novel AI tool for the detection of sperm DNA fragmentation using the TUNEL assay as a reference standard. Our machine learning approach enables non-destructive assessment of sperm DNA integrity through the analysis of phase-contrast and bright-field microscopy images alone. With a sensitivity of 70\%, the system demonstrates significant potential for improving sperm selection in assisted reproductive technologies by ensuring that only sperm with intact DNA are utilised.\\

The ability to perform rapid, non-invasive sperm assessment represents a transformative advancement in reproductive medicine, offering both increased efficiency and accessibility. By optimising sperm selection based on DNA integrity, this approach has the potential to enhance the success rates of ART procedures, ultimately improving outcomes for individuals and couples struggling with infertility.
Future work will focus on expanding the training dataset to further refine model performance and implementing the system in clinical settings to validate its impact on fertility treatment outcomes.

\section*{Acknowledgement}
The authors would like to acknowledge and thank the participating laboratories, BioART Fertility Centre and the Reproductive Biology Lab at the Steve Biko Academic Hospital, for the participation and support of this work.



\appendix
    \section{Data Examples}\label{sec:data_example}
    \subsection{Comparison of fragmented, unfragmented, and null-labeled sperm under different imaging techniques.}
    The following presents examples of images for each of the class types and illumination techniques.
    \begin{figure*}[t!]
        \centering
        \begin{subfigure}[b]{0.3\textwidth}
            \centering
            \includegraphics[width=\textwidth]{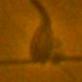}
            \caption{Fragmented - Phase-Contrast}
            \label{fig:frag_pc}
        \end{subfigure}
        \hfill
        \begin{subfigure}[b]{0.3\textwidth}
            \centering
            \includegraphics[width=\textwidth]{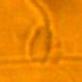}
            \caption{Fragmented - Bright-Field}
            \label{fig:frag_bf}
        \end{subfigure}
        \hfill
        \begin{subfigure}[b]{0.3\textwidth}
            \centering
            \includegraphics[width=\textwidth]{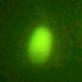}
            \caption{Fragmented - Fluorescent}
            \label{fig:frag_fl}
        \end{subfigure}
        
        \vspace{1em} 

        \begin{subfigure}[b]{0.3\textwidth}
            \centering
            \includegraphics[width=\textwidth]{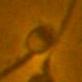}
            \caption{Unfragmented - Phase-Contrast}
            \label{fig:unfrag_pc}
        \end{subfigure}
        \hfill
        \begin{subfigure}[b]{0.3\textwidth}
            \centering
            \includegraphics[width=\textwidth]{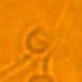}
            \caption{Unfragmented - Bright-Field}
            \label{fig:unfrag_bf}
        \end{subfigure}
        \hfill
        \begin{subfigure}[b]{0.3\textwidth}
            \centering
            \includegraphics[width=\textwidth]{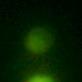}
            \caption{Unfragmented - Fluorescent}
            \label{fig:unfrag_fl}
        \end{subfigure}

        \vspace{1em} 

        \begin{subfigure}[b]{0.3\textwidth}
            \centering
            \includegraphics[width=\textwidth]{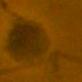}
            \caption{Null - Phase-Contrast}
            \label{fig:null_pc}
        \end{subfigure}
        \hfill
        \begin{subfigure}[b]{0.3\textwidth}
            \centering
            \includegraphics[width=\textwidth]{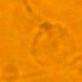}
            \caption{Null - Bright-Field}
            \label{fig:null_bf}
        \end{subfigure}
        \hfill
        \begin{subfigure}[b]{0.3\textwidth}
            \centering
            \includegraphics[width=\textwidth]{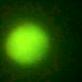}
            \caption{Null - Fluorescent}
            \label{fig:null_fl}
        \end{subfigure}

        \caption{Comparison of fragmented, unfragmented, and null-labeled sperm under different imaging techniques.}
        \label{fig:sample_grid}
    \end{figure*}
        
\printcredits

\bibliographystyle{model1-num-names}

\bibliography{cas-refs.bib}





\end{document}